# Measuring Economic Policy Uncertainty Using an Unsupervised Word Embedding-based Method

Fatemeh Kaveh-Yazdy[*]

Department of Computer Engineering,
Yazd University, Yazd, Iran
fkavehy@stu.yazd.ac.ir
(ORCID ID: 0000-0003-3559-5510)

Sajjad Zarifzadeh

Department of Computer Engineering,
Yazd University, Yazd, Iran
szarifzadeh@yazd.ac.ir

**Abstract:**

Economic Policy Uncertainty (EPU) is a critical indicator in economic studies, while it can be used to forecast a recession. Under higher levels of uncertainty, firms' owners cut their investment, which leads to a longer post-recession recovery. EPU index is computed by counting news articles containing pre-defined keywords related to policy-making and economy and convey uncertainty. Unfortunately, this method is sensitive to the original keyword set, its richness, and the news coverage. Thus, reproducing its results for different countries is challenging. In this paper, we propose an unsupervised text mining method that uses word-embedding representation space to select relevant keywords. This method is not strictly sensitive to the semantic similarity threshold applied to the word embedding vectors and does not require a pre-defined dictionary. Our experiments using a massive repository of Persian news show that the EPU series computed by the proposed method precisely follows major events affecting Iran's economy and is compatible with the World Uncertainty Index (WUI) of Iran.

**Highlights:**

- EPU is a critical indicator in economic studies to predict future investments, the unemployment rate, and recessions.
- An unsupervised, Omni-language method that uses word embedding representations is proposed.
- Five EPU measures are computed for Iran from 2015 to 2018 using a massive repository of 10M news articles.
- The proposed EPU measure follows the changes in Iran's policy tightly.

---

[*] Corresponding author at: Department of Computer Engineering, Yazd University, Yazd, Iran. PO Box: 98195 – 741.
E-mail addresses: fkavehy@stu.yazd.ac.ir (*F. Kaveh-Yazdy*), szarifzadeh@yazd.ac.ir (*S. Zarifzadeh*).

# 1 Introduction

While financial relations in countries are regulated by governments, uncertainty imposed by the policy made can influence the various aspects of the economy such as investment, income, tax systems and public services. The uncertainty which is implied by changes in economic policies made by governments is called economic policy uncertainty (EPU). Economic uncertainty can be defined as the public's inability to predict the outcomes of their decisions under new policies applied to financial relations. Economic policy uncertainty (EPU) due to its impacts on economic growth has been widely studied by the economic community. When the EPU is high, firms' owners and households postpone their irreversible investors and hold their cash, a.k.a. wait-and-see policy [1]. Studying EPU is proliferated after the 2008 global financial crisis (GFC) and takes leaps in recession prediction [2], unemployment rate tracking [3], investment [4], post-recession recovery [5].

The real option theory indicated that the delay policy could be considered as an option for investors in situations with higher levels of uncertainty [6]. Moreover, the investment irreversibility and high capital adjusted costs that are higher under uncertain situations decrease the investment [7]. Economic policy uncertainty affects the different aspects of the economy through the two major channels, which are: (1) investment delay and (2) cash holding (cash flow inhibition). Decreasing the investment in firms and industries increases the unemployment rate [8] and production declining [9], [10]. On the other side, households and industries hold their cash by decreasing their expenditure, leading to lower demand for products and services. Indeed, putting a delay on investment and cash holding are associated behaviors [11]–[13] that affect consumption and production.

Uncertainty imposed by economic policy changes in a country does not only affect that country's economy but also disturbs the economies of the countries that are connected. For example, changes in economic uncertainty of the US influence the uncertainty and economic growth of Mexico [14] and China [16]–[18]. In addition to the local factors that boost the EPU (e.g., tax policy uncertainty, income changes), global crisis and geopolitical risks can affect the EPU of a country. Gulf Wars I and II [15], 9/11 attacks [1], Iraq war [16], terrorist attacks in Europe [17], 2008 global recession [3] as well as pandemics [18] are such events that have affected the economic uncertainty of different countries.

Among uncertainty measures, the economic policy uncertainty (EPU) index is widely studied. There are three groups of methods adopted measuring economic uncertainty. The first group uses economic/financial parameters, e.g. CBOE[1] volatility index (VIX)[2], to estimate the economic uncertainty. The second group uses text mining methods to extract the information from various textual resources, and the third group of methods uses miscellaneous sources of data to estimate the uncertainty. In this article, we intended to focus on text mining-based methods using news data to estimate the policy uncertainty realized by the investors. In this way, we review these methods and introduce a novel word embedding-based method designed to overcome the obstacles of the conventional methods.

The rest of this article is organized as follows. Section 2 surveys the economic policy uncertainty measures and demonstrate a hierarchy of methods. In section 3, the news repository is introduced, and the data preprocessing steps are enumerated, then the proposed word embedding-based method is presented. Section 3 is finalized by reviewing the baseline EPU indices implemented to compare with the proposed index. Section 4 discusses the results of our comparative experiments as well as correlation analysis. This article is concluded in section 5.

---

[1] Chicago Board Options Exchange (CBOE)
[2] https://www.cboe.com/tradable_products/vix/

# 2 Economic Policy Uncertainty Measures

Economic uncertainty measures can be divided into three categories: (1) financial measures, (2) textual measures, and (3) miscellaneous methods (*ref.* Figure 1 for hierarchy of methods). The first class ($C_1$) adopted financial parameters to estimate the implied economic policy. These measures are constructed based on two major ideas, (A) measuring volatility in the market and (B) estimating the unpredictability of economic parameters. The idea behind the methods defined based on the volatility in the stock market ($C_1.A$ methods) comes from the results of Bloom [7]. He showed that the realized uncertainty is negatively associated with productivity growth which can be estimated by taking the stock market volatility into account. The volatility index (VIX for short), which has been computed since 1992 by the Chicago Board Options Exchange (CBOE) is an uncertainty measure that belongs to $C_1.A$ group. VIX measure, which is also known as the fear index, is widely used in economic uncertainty analysis [19]–[21] and stock price prediction.

The second group in the first class ($C_1.B$) includes measures that are constructed by comparing the expected values of financial parameters with their real values. Intuitively, the higher the uncertainty, the more fluctuate the economy is, and analysts' forecasts are more erroneous. This group uses two types of data that are (1) forecasters' stock market predictions ($C_1.B.D_1$) and (2) surveys ($C_1.B.D_2$). Lahiri and Sheng [22] (from the $C_1.B.D_1$ group) examined the disagreement between forecasts of a group of analysts and actual values and showed the differences could be used as an uncertainty proxy. Survey-based methods are linked with measures defined based on the forecasters' errors. Measures in this group ($C_1.B.D_2$) compare the expected and actual values of the macroeconomics parameters (such as GDP and inflation) that are collected via surveys. Boards of professional analysts who collaborate with central banks present their short-/long-run predictions, and the dispersion between values are used as an uncertainty proxy. For example, European Central Bank (ECB) has adopted a panel of one hundred experts to provide long-term predictions of GDP, inflation, and the unemployment rate in Europe which is called ECB survey Professional Forecasters (ECB-SPF). Poncela and Senra [28] constructed an uncertainty measure based on ECB-SPF, which reduces the dimension of the panel data using principal component analysis. Finally, a panel regression applied to the transformed data predicts the uncertainty.

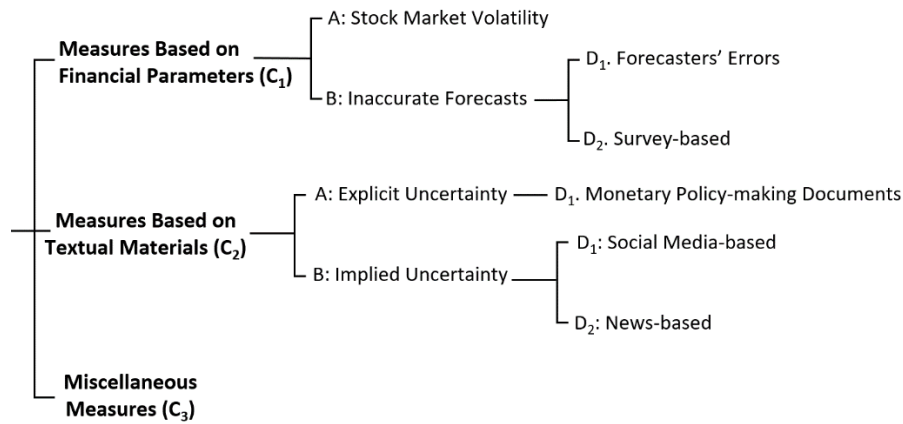

*Figure 1: Categories of economic uncertainty and economic policy uncertainty measures grouped based on type of their input data.*

The second class of methods includes measures that are constructed using textual data. These measures can be divided into two groups with respect to their source of data. The first group (i.e., $C_2.A.D_1$) tries to measure the uncertainty directly from monetary policy-making documents. These methods track the changes made in economic policy reflected in the documents, such as minutes of decision boards in the Central Banks. In this way, Lee et al. [23] collected minutes of the Bank of Korea (BoK) monetary policy board and extracted two uncertainty indices. These indices measure the tone of sentences using an economic field-specific dictionary and a market analysis approach. Shifts in the dictionary-based measure together with the market-based index demonstrate the uncertainty in the policy-making of the Bank of Korea. Hansen et al. [24] investigated the relationship between the contents of the minutes of central banks and uncertainty. They analyzed minutes of the Federal Reserve, the Bank of England, and the European Central Bank using LDA topic modeling and found that the EPU is linked with the terms that represented the fiscal and risk topics.

The second group of measures in class $C_2$ (that are $C_2.B$) uses the textual information collected from social media ($C_2.B.D_1$) and news articles ($C_2.B.D_2$). Social media posts reveal the uncertainty realized by the households and public communities [18]. Baker et al. [25] introduced Twitter-based Economic Uncertainty (TEU). They adopted this measure together with news-based and financial uncertainty indices to track the impacts of the COVID-19 pandemics in the US and UK. Results showed diverse behavior among the measures, which confirm the variety of aspects they were taken into account. Becerra and Sagner [26] conducted an investigation to find the impact of the COVID-19 pandemic on the uncertainty level in Chile using the EPU index. Their results are consistent with that presented by Baker et al. [25] and show uncertainty peak in a few weeks after the disease outbreaks in Chile.

While owners of the firms and households get information from news sources, Baker et al. [1] inferred that the best source for measuring uncertainty realized by investors is news articles. In fact, most investors look at the market and economy through the lens of news and analytical reports.

Baker's et al. [1] measure as a news-based index is widely used to estimate the economic policy uncertainty. They searched the archive of ten US newspapers and selected the articles that included terms "uncertainty" or "uncertain"; "Economy" or "Economic," and a set of terms conveying the policy-making meaning, e.g., "legislation," "Congress," and "regulation." The raw number of articles that pass the conditions in each newspaper over each month is normalized by the total number of articles in the corresponding newspaper and month. The normalized monthly series are standardized to unit deviation and then divided by the monthly averages across newspapers. The normalized series are scaled to a mean of 100. Baker et al. [1] computed the economic policy uncertainty index (henceforth, BDD[1] index) for twelve major economies, including the United States, China, Russia, Germany, and Japan[2]. Baker et al. [1] used English news article to estimate the uncertainty for non-Anglophone countries which increase the risk of under-estimation. Consequently, several research groups repeat the process by extracting articles from local newspapers and native keywords, e.g., Huang and Luk [27] computed the BBD index using ten major Chinese newspapers published in mainland China. Arbatli et al. [28], Ferreira et al. [29], and Ghirelli et al. [30] constructed native BBD indices for Japan, Brazil, and Spain, respectively. BBD index relies on the coverage of keywords and requires human supervision for keyword selection.

Azqueta-Gavaldón et al. [31] tried to overcome the supervision requirements in measuring uncertainty by adopting unsupervised topic modeling. They cleaned the documents, removed stopwords, and applied a Latent Dirichlet Allocation (LDA) method. The LDA is a generative model which groups observations, that are words, in unobserved groups, namely topics. In this model, each document can be viewed as a mixture of a small set of topics. Azqueta-Gavaldón et al. [31] merged topics under the eight sectors of uncertainty initially indicated by Baker et al. [1].

---

[1] BDD comes from initials of Baker, Bloom, and Davis
[2] Available at www.policyuncertainty.com

Although their method is unsupervised, aggregating the topics under the umbrellas of the sectors requires human supervision. Yono et al. [32] combined the supervised LDA (sLDA) method with the VIX index and constructed a fully unsupervised EPU index. Their model used the VIX index as a supervision signal and trained it to forecast the VIX measure by taking into account the distribution of the words in topics.

Tobbak et al. [33] labeled a set of documents considering economic uncertainty and a support vector machine (SVM) classifier to label documents with EPU-relevant/not-EPU-relevant tags. For each document, the likelihood of being EPU-relevant is extracted, and documents with higher values are forwarded to an active learning algorithm. The active learner optimizes the decision boundaries of the SVM classifier. Keith et al. [34] sampled 2531 labeled documents from Baker's et al. [1] corpus and divided it into two sets of 1844 documented from 1985-2007 as the training set and 687 documents from 2007-2012 as the test set. A dictionary of unique terms used in the cleaned version of documents generated to project the documents set into a multi-feature representation space. The bag-of-word representation is used by a logistic regression classifier to assign the EPU-relevant/not-EPU-relevant labels. Keith et al. [34] examined the combination of a classifier with a word embedding-based representation of the documents, which was surpassed by the bag-of-word-based representation classifier. Tsai and Wang [35] expanded the economic dictionary developed by Loughran and Mcdonald [36] using the Continuous Bag-of-Words (CBOW) model. They find the top-20 most similar words and add them to the dictionary of Baker et al. [1].

## 3 Materials and Methods

### 3.1 Data

We collect logs of the news service of the Parsijoo Persian search engine[1] from January 2015 to December 2018. This textual repository contains 10,532,325 Persian news articles published by 62 Persian news agencies. These articles may include multimedia content such as photos, videos, and voices. This repository can be seen as an integrated all-in-one news resource for text mining. In this log, for each article, a minimum set of properties, i.e., ID, title, body text, topic, and their publishing date/time are stored. Titles and body texts of articles are merged, multimedia contents are removed, and words are stemmed.

### 3.2 Word Embedding-based Uncertainty Index

A dictionary, similar to the one in Loughran and Mcdonald [36], has not been authored for Persian financial/economic studies yet. Moreover, recent studies showed that word embedding-based keyword expansion is more successful than that of relying on experts' knowledge. Therefore, we decide to use word embedding-based representation for tagging relevant documents. Applying restriction on the number of semantically similar terms used in document selection (same as Tsai and Wang [35]) or setting a threshold on semantic similarity (similar to Rekabsaz et al. [37] and Theil et al. [38]) raise the same challenge which is determining the threshold for expanding the concepts of Economy, Policy, and Uncertainty. Strict thresholds omit relevant documents that address uncertainty with mild terms, and relaxed thresholds let the irrelevant documents to be retrieved and counted. To overcome this challenge, we select a relaxed threshold and then scale their impact. We assign weights to each document based on the cosine similarity between its words and the three seed words, i.e., Economy, Policy, and Uncertainty, henceforth, called major concepts. In this way, we decrease our similarity threshold to values lower than that specified by Rekabsaz et al. [39] to ensure that all words that can address the meaning represented by these major concepts are selected. Then, we represent each document in a simple tri-axis system and scale their impact by a score.

---

[1] Parsijoo news service can be accessed at news.parsijoo.ir

Suppose that corpus, $\mathbb{C}$, is used to train a word embedding representation of a set of words called the dictionary, i.e., $\mathfrak{D}$.

**Definition 1** (Major Concepts): Economy, Policy, and Uncertainty are major concepts denoted by $E$, $P$, and $U$, respectively. These concepts are represented by the words "economy", "policy", and "uncertainty", which are the members of the dictionary[1].

**Definition 2** (Vectors of Major Concepts): Suppose that there exists a word embedding representation of words that is trained using news articles. Let $\vec{V_E}$, $\vec{V_P}$ and $\vec{V_U}$ be the embedding vectors of major concepts $E$, $P$, and $U$, respectively.

**Definition 3** (Cosine Similarity): Cosine similarity is a metric used to measure the similarity between two vectors based on the cosine of the angle between them. Let $\vec{V_x} = (x_1, x_2, \dots, x_n)$ and $\vec{V_y} = (y_1, y_2, \dots, y_n)$ be vectors with $n$ elements, and $\theta$ be the angle between them. The cosine similarity between two vectors is computed as:

$$Sim(\vec{V_x}, \vec{V_y}) = Cosine(\theta) = \frac{\vec{V_x} \cdot \vec{V_y}}{\|\vec{V_x}\|\|\vec{V_y}\|} = \frac{\sum_{i=1}^{n} x_i y_i}{\sqrt[2]{\sum_{i=1}^{n} x_i^2} \sqrt[2]{\sum_{i=1}^{n} y_i^2}} \tag{1}$$

**Definition 4** (Semantically Similar Words): Let $w_i$ and $w_j$ be words such that $w_i, w_j \in \mathfrak{D}$. This pair is called semantically similar words, if the cosine similarity of their embedding vectors satisfies a minimum similarity threshold denoted by $T_{min}$, as:

$$Sim(\vec{V}_{w_i}, \vec{V}_{w_j}) \geq T_{min} \tag{2}$$

**Definition 5** (Nearest Words): Let $D$ be a news document represented as a set of words (a.k.a. bag-of-word), i.e., $D = \{w_1, w_2, \dots, w_m\}, \forall w_i \in \mathfrak{D}$. There exist three words $w_i$, $w_i$ and $w_i$ such that

$$w_i = \text{argmax } Sim(\vec{V}_{w_i}, \vec{V_E}) \tag{3}$$

$$w_j = \text{argmax } Sim(\vec{V}_{w_j}, \vec{V_P}) \tag{4}$$

$$w_k = \text{argmax } Sim(\vec{V}_{w_k}, \vec{V_U}) \tag{5}$$

**Definition 6** (Semantically Similar Nearest Words): Nearest words of each document are indicated as semantically similar words, if their similarity values to their corresponding major concept satisfy the minimum threshold:

$$Sim(\vec{V}_{w_i}, \vec{V_E}) \geq T_{min} \tag{6}$$

$$Sim(\vec{V}_{w_j}, \vec{V_P}) \geq T_{min} \tag{7}$$

$$Sim(\vec{V}_{w_k}, \vec{V_U}) \geq T_{min} \tag{8}$$

---

[1] We use major concepts and their word representations, interchangeably.

For the sake of easiness, the similarity values of semantically similar nearest words are addressed using three formal variables, i.e., $\alpha$, $\beta$, and $\gamma$. If a similarity measure does not satisfy the threshold condition, we set the formal variables to zero, as follows:

$$\alpha = \begin{cases} Sim(\vec{V}_{w_i}, \vec{V}_E) & if\ Sim(\vec{V}_{w_i}, \vec{V}_E) \geq T_{min} \\ 0 & if\ Sim(\vec{V}_{w_i}, \vec{V}_E) < T_{min} \end{cases} \quad (9)$$

$$\beta = \begin{cases} Sim(\vec{V}_{w_j}, \vec{V}_P) & if\ Sim(\vec{V}_{w_j}, \vec{V}_P) \geq T_{min} \\ 0 & if\ Sim(\vec{V}_{w_j}, \vec{V}_P) < T_{min} \end{cases} \quad (10)$$

$$\gamma = \begin{cases} Sim(\vec{V}_{w_k}, \vec{V}_U) & if\ Sim(\vec{V}_{w_k}, \vec{V}_U) \geq T_{min} \\ 0 & if\ Sim(\vec{V}_{w_k}, \vec{V}_U) < T_{min} \end{cases} \quad (11)$$

For each news document, there exist three features representing this document in the semantic space, which embed all dictionary words and major concepts. The values of $\alpha$, $\beta$, and $\gamma$ belong to the most similar terms that satisfy the minimum similarity threshold. While the similarity threshold is not determined as high as the thresholds used in similar methods, documents covering uncertainty with mild terms would be selected. To avoid overestimation, we prefer to select documents covering all facets of EPU at the medium level, instead of documents that cover one facet at its maximum level. For example, suppose that document ($A$) includes the words "uncertainty", "economy" and "minster", and document ($B$) contains "uncertainty", "financial", and "law-making". However, word "uncertainty" and "economy" are fully matched with major concepts in ($A$) and only "uncertainty" is matched in document ($B$). However, the distance between the "minister" in the document ($A$) to "policy" is more than the distances between "financial" and "law-making" to "economy" and "policy" concepts. Thus, we propose a triangular document representation that considers the impacts of all major concepts together.

**Definition 7** (Tri-axial Representation System): The proposed representation system is a non-orthogonal tri-axis system containing three axes that meet in the origin point with an equal angle of $\frac{2\pi}{3}$ between them. This representation system can be represented in a 2D Cartesian coordinate system. Figure 2 shows the proposed representation system.

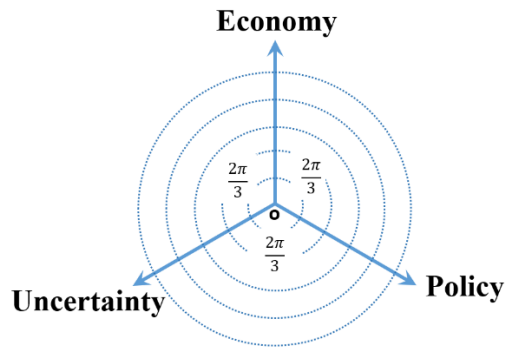

*Figure 2: The proposed tri-axis document representation system.*

**Definition 8** (Triangular Document Representation): Let $D$ be a news document. The similarity values of $D$ to the major concepts, i.e., $\alpha$, $\beta$, and $\gamma$, are mapped on axes of the tri-axial system. The triangle constructed by connecting the mapped points on axes is the triangular representation of $D$, as illustrated in Figure 3.

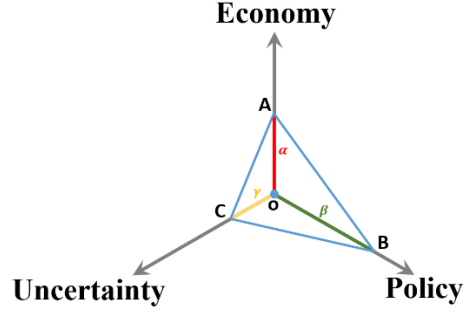

*Figure 3: Triangular representation of document D given similarity values of α, β and γ.*

The length of the triangle's sides are computed as follows[1],

$$L_{AB} = \sqrt[2]{\alpha^2 + \beta^2 + \alpha\beta} \tag{12}$$

$$L_{AC} = \sqrt[2]{\alpha^2 + \gamma^2 + \alpha\gamma} \tag{13}$$

$$L_{BC} = \sqrt[2]{\beta^2 + \gamma^2 + \beta\gamma} \tag{14}$$

**Definition 9** (EPU Score): Let $D$ be a document represented in the tri-axial representation system. The area under the triangle named EPU score and can be calculated based on Heron's formula as follows,

$$\text{EPU Score} = \sqrt[2]{S(S - L_{AB})(S - L_{AC})(S - L_{BC})} \tag{15}$$

where $S$ is $\frac{L_{AB} + L_{BC} + L_{AC}}{2}$.

We should note that the EPU score of a document that is received zero for one of its similarity measures is set to zero. In our implementation, the semantic similarity threshold is set to 0.5, and the EPU scores of documents are summed in each month and normalized by dividing them by the total number of articles in the corresponding month. Finally, normalized numbers are standardized to unit standard deviation.

## 3.3 Baseline Uncertainty Index Construction

We selected four baseline EPU indices to compare with our method: (1) Baker's et al. [1] EPU index, (2) Azqueta-Gavaldón [40] LDA-based EPU index, (3) Braun's [41] EPU Index, and (4) World Uncertainty Index of Iran computed by Bloom et al. [42] and presented by the Federal Reserve Bank of St. Louis. As a test case study, these EPU indices are implemented and tested using Persian news posts to estimate Iran's EPU index.

We translated Baker's et al. [1] core words, i.e., "economy", "policy", and "uncertainty" using online English-to-Persian Abadis dictionary[2]. Then, the Persian terms are expanded using Farsnet[3] Persian ontology. Documents containing words from all three groups are retrieved and counted. We should note that the news service log, as we received, does not separate news sources; thus, our EPU index has only one series of monthly counts. Monthly counts are normalized using the total number of articles in each month and standardized to unit standard deviation with zero mean values.

---

[1] Details of computations are addressed in Appendix A.
[2] Accessed at www.abadis.ir
[3] Farsnet is a Persian ontology which is designed based on the WordNet ontology and can be aligned to WordNet English ontology.

Azqueta-Gavaldón [40] uses economic news to estimate his LDA-based EPU index, which means he does not expand terms related to the economy. Instead, he uses the LDA method to group documents based on the statistical distribution of their terms in each topic. In the final step, he aggregates the topics under eight EPU groups. We select documents that are labeled with "Economy" and "Financial" in the Parsijoo news service. This set includes 1,164,102 news articles. The economic articles are stemmed, spell checked [43], and stop-words are removed. Words with extremely high frequency that appear in more than 60% of documents are omitted. In addition, we prune the dictionary by removing those words their frequencies are less than 9000. The remaining set of words includes 2185 items. In the next step, documents are represented in a vector-based representation. Each element in $i^{th}$ row and $j^{th}$ column of this representation is the total number of times the word $j$ appears in $i^{th}$ document. The size of the generated matrix is $1,164,102 \times 2185$. LDA model is a hierarchical Bayesian model that uses the vector representation of documents to model documents in a space with a smaller number of dimensions with respect to the probability of co-occurred words in the same topics. It is worth mentioning that the number of topics must be determined before modeling. There are various measures to decide about the number of topics in LDA, which are studied and compared by Roder et al. [44]. We decide about the number of topics based on the coherence score of the topics, $C_v$.

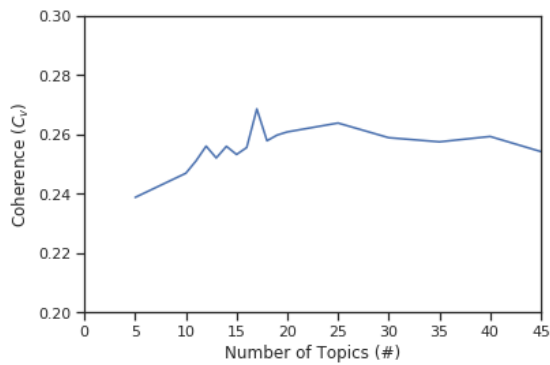

*Figure 4: Number of topics and topic coherence scores of LDA models.*

Figure 4 shows the coherence scores of the LDA models with a given number of topics. With respect to the values of the scores, we trained the model with 17 topics. This model is used to label all economic news articles. Each topic is demonstrated by its top $k$ (namely $k = 40$) keywords. Among the discovered topics, we identify the set of topics that have uncertainty-related terms in their top 40 keywords and label them as uncertainty topics. Documents labeled with uncertainty topics are counted across the corpus. The series is normalized by the total number of articles published in the corresponding month and standardized to unit standard deviation.

The third benchmark EPU index selected for comparison studies is Braun's [41] EPU Index. Braun [41] counts the number of documents that demonstrated the uncertainty concepts and multiply count into the number of economic news articles containing policy-making keywords. The finalized series is standardized to unit standard deviation.

Bloom et al. [42] compute the quarterly World Uncertainty Index; thus, its raw values cannot be compared with the proposed EPU index. The WUI is computed using the normalized frequency of the word "uncertainty" in the quarterly Economist Intelligence Unit (EIU) country reports. We decompose the WUI to the elements that can be counted in Persian news articles. In other words, we find a linear/non-linear system of counts of Persian terms that can be used to predict the values of WUI for the corresponding season and use this model to forecast monthly WUI values. Figure 5 shows the schema of the prediction model.

In the first step, documents of each quarter are stacked to generate $4 \times 4$ bulks of articles, and the next step would be extracting vectors of terms from bulks. HaCohen-Kerner et al. [45] investigations showed that bag-of-word text classification methods could be significantly benefitted from stop-word removal. Moreover, results of Kaveh-Yazdy et al. studies [46], [47] showed that removing most frequent words (that are appeared in more than $p$ % of documents) as well as stop-word removal in Persian news mining applications produce better results. Therefore, most frequent words (words that appeared in more than 60% of articles) and stop-words are removed before vectorization. The bag-of-word representation of the corpus is a matrix of 16 rows (eq. the number of bulks) and 5975 columns (eq. the number of words), i.e., $X_{16 \times 5975}$. This matrix is used to estimate the quarterly WUI of 2015-2018 that is $Y_{16 \times 1}$. Estimating $Y$ using inputted $X$ (as a regression problem) would be challenging if the number of predictor is more than the number of observations.

There are various methods to solve regression problems with more predictors than observations. Among these methods, Partial Least Squares Regression (PLS Regression) [48] is particularly appropriate for this problem. PLS regression projects matrices $X$ and $Y$ to a new space represented by components called latent structures and then, solves the new problem. This method is robust to multicollinearity among features of $X$ [49]. Collinearity might happen in vectors representing text mining problems (e.g., Ye et al. [50] addressed this issue), which means a predictor variable has a linear association with another predictor. When there exist multiple pairs of collinear variables, the phenomenon is called multicollinearity. The PLS regression requires the number of latent structures (a.k.a. components) to be determined before estimating $Y$; thus, we run the algorithm with a different number of components and select the best state with respect to the estimation's mean squared error (MSE).

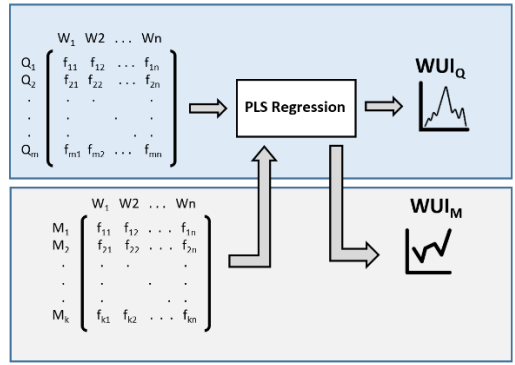

*Figure 5: Schema of the prediction model which forecasts monthly WUI values using PLS regression trained on quarterly WUI data.*

Figure 6 shows the MSE values of the PLS regression for a given number of components. Thus, the best number of components for PLS regression is 15. In the next step, the number of times words occurred in the bulks of the news belongs to each month are summed up to generate the $X'_{48 \times 5975}$ Matrix. The trained model is fed with $X'_{48 \times 5975}$ and predicts the $Y'_{48 \times 1}$, i.e. estimated monthly WUI. The estimated WUI is standardized to unit standard deviation and adopted in comparison with the proposed method.

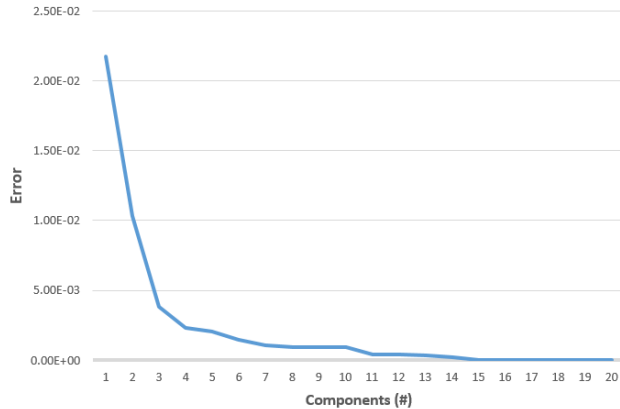

*Figure 6: Mean Square Error (MSE) of PLS regression for a given number of components.*

# 4  Results and Discussion

In this section, we compare the proposed EPU index with respect to its ability to show historical events and its application in studying social data analysis. Figure 7 illustrates the comparative chart of the baseline EPU indices as well as the proposed EPU index. As it can be seen, the proposed index follows the WUI in major spikes; however, from January 2015 to April 2016, the proposed index is less fluctuating than the WUI. In this period, other indices that are directly generated from Persian news have similar behavior. This graph indicates that Iran's EPU in recent years has been linked to the events that happened around Iran and the P5+1[1] agreement (a.k.a. The Joint Comprehensive Plan of Action or JCPOA, in brief). An interesting phenomenon is the difference between WUI and other indices. In almost all cases, the WUI dramatically spikes, while other indices change slower.

In this paper, we should note that the WUI is an estimated series of Bloom's et al. [42] WUI that is computed using a PLS regression model on Persian news. Bloom et al. [42] extracted their index by mining Economist Intelligence Unit (EIU) country reports, which means that this index is the manifestation of Iran's economy from EUI's experts' perspective and others show the national perception. It can be inferred that Iran's economy is more resilient than expected, and WUI imprecisely models its socio-economical responses to these events. While WUI spikes coincide with most critical events in the studied period; however, after US withdrawal from JCPOA, WUI is not notably increased. In this event, BBD, Braun, and Azqueta-Gavaldón indices show a lower level of uncertainty, which is not reasonable. In addition, the Switzerland statement on key parameters of JCPOA (henceforth, Switzerland statement) was announced in April 2015. During the negotiations which lead to this statement, Iran and P5+1 agreed on the framework and parameters of the agreement. In other words, the Switzerland statement was a preliminary step before the agreement. Although it is expected that the Switzerland statement reduces the uncertainty, the values of BBD, Azqueta-Gavaldón, and Braun indices show incremental behaviors. Indeed, WUI and the proposed index follow the major events that affect the economy, even though the values of WUI show drastic changes. Table 1 shows the correlation matrix of the baseline indices and the proposed index. The highest correlation between the proposed index and the estimated WUI confirms findings regarding their similar behavior. In Table 1, the correlation of the BBD, Barun, and Azqueta-Gavaldón are closer together. We use hierarchical clustering to group indices with respect to the values of series.

---

[1] The five permanent members of the United Nations Security Council—China, France, Russia, United Kingdom, United States—plus Germany.

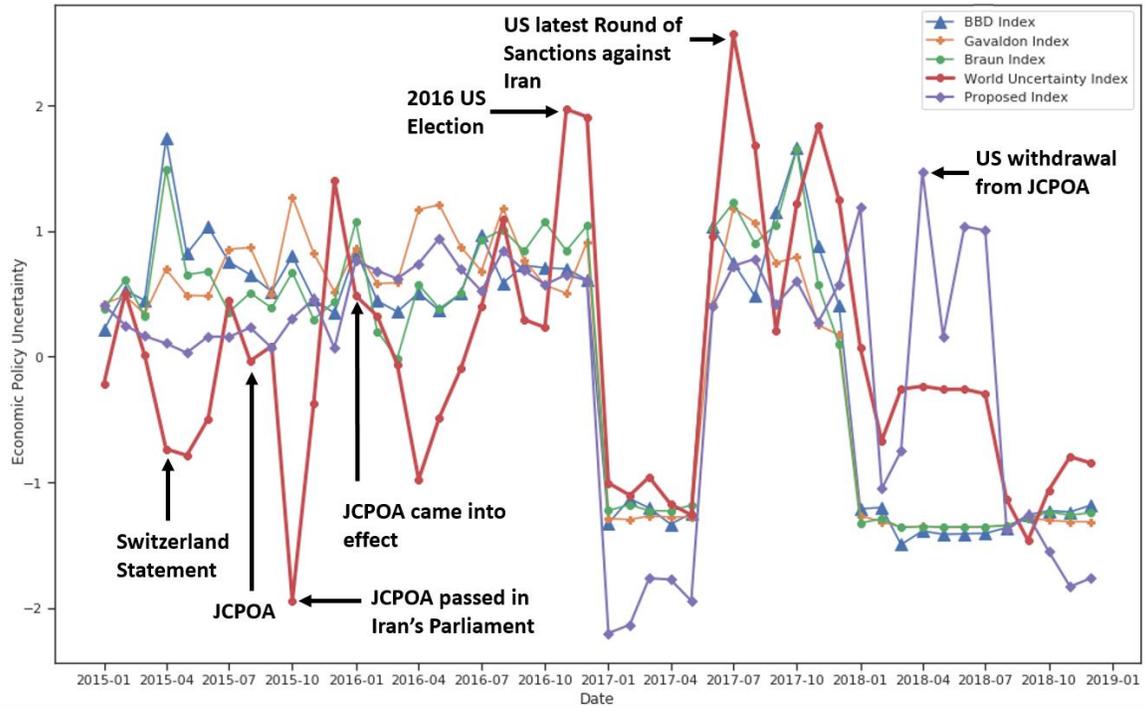

*Figure 7: Comparison between (1) Baker's et al.* [1] *EPU index, (2) Azqueta-Gavaldón* [40] *LDA-based EPU index, (3) Braun's* [41] *EPU Index, and (4) estimated World Uncertainty Index of Iran computed by Bloom et al.* [42] *and the proposed EPU Index. All series are standardized to unit standard deviation.*

**Figure 8** shows the dendrogram representation of the indices, which are hierarchically clustered. As it is expected, BBD, Barun and Azqueta-Gavaldón are grouped together, and the proposed index is merged with their cluster; the last index, which is more different, is the estimated WUI. An important question remaining is whether the first group, including BBD, Barun, and Azqueta-Gavaldón indices, is more precise than the second one.

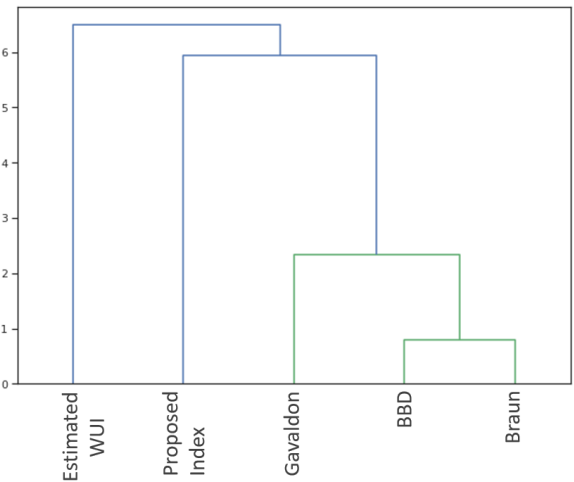

*Figure 8: The dendrogram representation of the hierarchical clustering of EPU indices.*

*Table 1: Correlation matrix of EPU Indices.*

| # | Index | (1) | (2) | (3) | (4) | (5) |
|---|---|---|---|---|---|---|
| 1 | Baker's et al. [1] | 1.0000 | | | | |
| 2 | Braun's [41] | 0.9935 | 1.0000 | | | |
| 3 | Azqueta-Gavaldón [40] | 0.9339 | 0.9434 | 1.0000 | | |
| 4 | Estimated WUI [42] | 0.5194 | 0.4720 | 0.5081 | 1.0000 | |
| 5 | Proposed Index | 0.5723 | 0.5667 | 0.6320 | 0.5602 | 1.0000 |

To answer this question, we conduct an experiment to investigate the properness of the indices with respect to their ability to follow the social responses to different levels of uncertainty in Iran. Antonakakis and Gupta [51], Vandoros et al. [52], and de Bruin et al. [53] studied the association between uncertainty and suicide cases in the US, England, and 17 other countries, respectively. These studies showed that the uncertainty value and the number of suicide cases are positively correlated. They confirmed that the number of suicide committing grows as the uncertainty increases. Therefore, the correlation analysis of the suicide series and EPU indices can reveal which index is a better representative measure to study the uncertainty. In this way, we extract all articles which disclose a suicide case. Whereas a suicide case may be covered by different news sources, we used a locality sensitive hashing method proposed by Montanari et al. [54] to detect near-duplicate news and avoid over-counting.

After removing near-duplicate articles, we count the number of suicide cases reported in the articles. Then, the monthly series of suicide cases is used in correlative studies to show which index has a higher correlation. Table 2 shows statistically significant positive correlation values between all indices and the series of suicide cases.

Our findings confirm the reported results of Antonakakis and Gupta [51], Vandoros et al. [52], and de Bruin et al. [53], considering the positive correlation between suicide cases and EPU indices. Furthermore, the correlation between the proposed index is higher than the other indices. Our social findings, as well as the ability of the proposed index to follow the notable events in Iran's foreign policy, show its reliability and precision. Besides, it is compatible with the estimated Bloom et al. [42] World Uncertainty Index.

*Table 2: Correlation of suicide series and EPU Indices.*

| # | Index | Correlation | P-value level |
|---|---|---|---|
| 1 | Baker's et al. [1] | 0.4123 | < 1% |
| 2 | Braun's [41] | 0.3833 | < 1% |
| 3 | Azqueta-Gavaldón [40] | 0.4082 | < 1% |
| 4 | Estimated WUI [42] | 0.4049 | < 1% |
| 5 | The Proposed Index | 0.5771 | < 1% |

## 5 Conclusion and Future Works

Economic uncertainty delays the investment plans directly and increases the unemployment rate, and triggers the cash holding. Results of past research works confirmed that uncertainty shocks lead to sharp recessions [7], [56]–[58], and Ercolani and Natoli [2] showed that the recessions could be predicted using the uncertainty level. The economic policy uncertainty is the uncertainty realized by the investors. The policy uncertainty can be measured indirectly from the uncertainty reflected in the financial documents, or the uncertainty implied through the news. There are four groups of methods used to measure economic policy uncertainty.

The first group of the methods is keyword-based methods initially proposed by Baker et al. [1]. This group is widely used and has shown its reliability. Baker et al. [1] used a pre-defined list of words representing the "Policy," "Economy" and "Uncertainty" concepts. Keyword selection procedure makes their method depend on the quality and coverage of the list. Moreover, it leads to compatible but different results for countries with languages other than English. The second group includes topic modeling methods that are proposed to overcome the supervision challenges of the keyword-based methods. Word embedding-based methods are proposed to expand the keyword sets of the Baker's et al. [1] method. The fourth group of methods uses Tf-IDF-based representations of the documents and labels them to adopt classifiers to annotate uncertainty-implied documents. All of these methods annotate documents and count the uncertainty-implication labels and count the

In this paper, we intended to review the text mining methods used for measuring the EPU index. Moreover, we propose a novel method that estimates the EPU index using automatically extracted keywords. This method is unsupervised and Omni-language. It lets different researchers reproduce similar results without being worried about the richness of the keyword set. Furthermore, this method does not need to determine precise similarity thresholds. Instead of selecting more similar keywords, it allows all documents which have at least one word higher than a relaxed threshold in each concept to be taken into account. However, the number of documents taken into account in this method is more than Baker et al. [12], but it can control each document's impact by its score.

Our comparative experiments showed that the proposed index coincides with major events in Iran's political space and follows the predicted response of the country's economy. This index is positively correlated with the World Uncertainty Index of Iran computed by Bloom et al. [42] and presented by the Federal Reserve Bank of St. Louis. In addition, we investigate the relation between the uncertainty level and the number of reported suicide cases, which shows a statistically significant positive correlation, i.e., higher than 0.5. An interesting point about Bloom's et al. [42] WUI is its ability to follow the major events; however, it shows sharper responses. We targeted studying different weighting schemes and feature selection methods to control the impact of documents in the overall score of each month for our future work. In addition, we found time windows included uncertainty-imposed terms expressing more than one change in the country's economic policy. The impacts of a new policy may exaggerate the effects of another altered policy. Therefore, it is reasonable to group the documents addressing a particular policy together. Then, the uncertainty indices of each group can be measured apart from others. A regression analysis of the panel data of uncertainty measures proxied by an exogenous variable such as stock market volatility can reveal different policies' impacts.

### Funding

This research did not receive any specific grant from funding agencies in the public, commercial, or not-for-profit sectors.

### CRediT authorship contribution statement

Will be provided after acceptance.

### Declaration of Competing Interest

The authors report no declarations of interest.

**Disclosure of conflicts of interest**

None.

**Acknowledgment**

None.

# Appendix A: Computational details of triangular representation

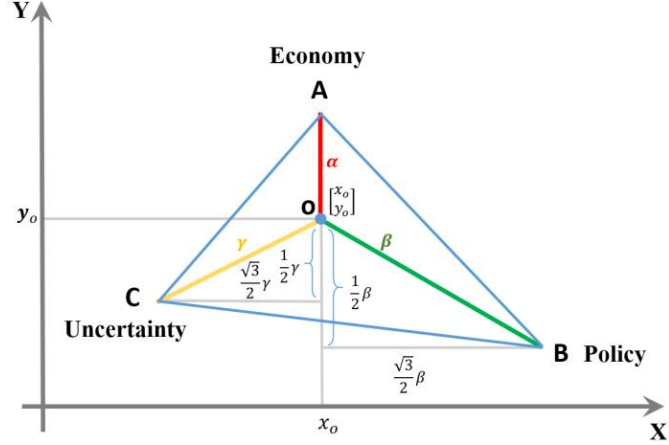

*Figure 9: Triangular representation of document D given similarity values of α, β and γ in 2D Cartesian space.*

Figure 9 shows the representation of a given document, i.e., $D$, in $2D$ Cartesian coordinate system. In this figure, the angles' positions are illustrated with respect to their offset from point $O$. According to this representation, the points $A$, $B$, and $C$ can be presented as follows,

$$A: \begin{cases} x_A = x_o \\ y_A = y_o + \alpha \end{cases} \tag{16}$$

$$B: \begin{cases} x_B = x_o + \dfrac{\sqrt{3}}{2}\beta \\ y_B = y_o - \dfrac{\beta}{2} \end{cases} \tag{17}$$

$$C: \begin{cases} x_C = x_o - \dfrac{\sqrt{3}}{2}\gamma \\ y_C = y_o - \dfrac{\gamma}{2} \end{cases} \tag{18}$$

The length of the triangle's sides equal to the Euclidean distances between points. The lengths of sides are calculated as,

$$\begin{aligned} L_{AB} &= \sqrt{(x_A - x_B)^2 + (y_A - y_B)^2} = \sqrt{\left(x_o - x_o - \frac{\sqrt{3}}{2}\beta\right)^2 + \left(y_o + \alpha - y_o + \frac{\beta}{2}\right)^2} \\ &= \sqrt{\frac{3}{4}\beta^2 + \alpha^2 + \frac{\beta^2}{4} + \alpha\beta} = \sqrt{\alpha^2 + \beta^2 + \alpha\beta} \end{aligned} \tag{19}$$

$$L_{AC} = \sqrt{(x_A - x_C)^2 + (y_A - y_C)^2} = \sqrt{\left(x_o - x_o - \frac{\sqrt{3}}{2}\gamma\right)^2 + \left(y_o + \alpha - y_o + \frac{\gamma}{2}\right)^2} \tag{20}$$

$$= \sqrt{\frac{3}{4}\gamma^2 + \alpha^2 + \frac{\gamma^2}{4} + \alpha\gamma} = \sqrt{\alpha^2 + \gamma^2 + \alpha\gamma}$$

$$L_{BC} = \sqrt{(x_B - x_C)^2 + (y_B - y_C)^2} = \sqrt{\left(x_o + \frac{\sqrt{3}}{2}\beta - x_o - \frac{\sqrt{3}}{2}\gamma\right)^2 + \left(y_o + \frac{\beta}{2} - y_o + \frac{\gamma}{2}\right)^2}$$

$$= \sqrt{\frac{3}{4}\beta^2 + \frac{3}{4}\gamma^2 - \frac{3}{2}\beta\gamma + \frac{\beta^2}{4} + \frac{\gamma^2}{4} + \frac{\beta\gamma}{2}} = \sqrt{\beta^2 + \gamma^2 + \beta\gamma}$$

(21)